# Transfer learning-based physics-informed convolutional neural network for simulating flow in porous media with time-varying controls

Jungang Chen

jungangc@tamu.edu

*Harold Vance Department of Petroleum Engineering, College of Engineering,*

*Texas A&M University, College Station, Texas, USA*

Eduardo Gildin, Ph.D.

*Professor, Harold Vance Department of Petroleum Engineering, College of Engineering,*

*Texas A&M University, College Station, Texas, USA*

and

John E. Killough, Ph.D.

*Professor(retired), Harold Vance Department of Petroleum Engineering, College of Engineering,*
*Texas A&M University, College Station, Texas, USA*

# Transfer learning-based physics-informed convolutional neural network for simulating flow in porous media with time-varying controls

**ABSTRACT**

A physics-informed convolutional neural network (PICNN) is proposed to simulate two phase flow in porous media with time-varying well controls. While most of PICNNs in existing literatures worked on parameter-to-state mapping, our proposed network parameterizes the solutions with time-varying controls to establish a control-to-state regression. Firstly, finite volume scheme is adopted to discretize flow equations and formulate loss function that respects mass conservation laws. Neumann boundary conditions are seamlessly incorporated into the semi-discretized equations so no additional loss term is needed. The network architecture comprises two parallel U-Net structures, with network inputs being well controls and outputs being the system states (e.g. oil pressure and water saturation). To capture the time-dependent relationship between inputs and outputs, the network is well designed to mimic discretized state space equations. We train the network progressively for every time step, enabling it to simultaneously predict oil pressure and water saturation at each timestep. After training the network for one timestep, we leverage transfer learning techniques to expedite the training process for subsequent time step. The proposed model is used to simulate oil-water porous flow scenarios with varying reservoir model dimensionality and aspects including computation efficiency and accuracy are compared against corresponding numerical approaches. The results underscore the potential of PICNN in effectively simulating systems with numerous grid blocks, as its computation time does not scale with model dimensionality. Furthermore, we assess the temporal error using 10 different testing controls with variation in magnitude and another 10 with higher alternation frequency with proposed control-to-state architecture. Our observations suggest the need for a more robust and reliable model when dealing with controls that exhibit significant variations in magnitude or frequency.

## 1  Introduction

Simulation of multiphase flow in heterogeneous porous media holds significant promise across various applications, including hydrocarbon recovery, greenhouse gas sequestration, and soil and groundwater remediation, among others [1, 2, 3, 4]. This is particularly true for two-phase flow simulations. We highlight that accurate yet efficient simulations of two-phase porous flow can lead to enhanced energy production, better environmental protection and more reliable engineering designs. Traditionally, the simulation of two-phase flow in porous media has relied upon numerical techniques such as finite difference, finite volume or finite element methods [3, 4, 5]. Following discretization of governing equations, Newton's method is normally performed to solve the discretized system and obtain high-fidelity solutions. This process is the most time-intensive part since it requires constructing Jacobian matrix with large dimensions, solving a linear system containing those Jacobians and iterating hundreds of times for every timestep until desired accuracy is attained. This becomes notoriously difficult when dealing with millions of grid cells and when thousands of forward runs are needed for tasks such as inverse modeling or optimization. Therefore, there's a pressing need for methods to reduce this computational complexity.

Deep learning models are increasingly employed in addressing applied mathematics and computational physics challenges, leveraging their capacity to learn complex patterns and relationships from data. Within the domain of subsurface flow applications, deep neural networks have been used for surrogate modelling [6, 7, 8, 9, 10, 11], uncertainty quantification [12, 13], and inverse modeling [14, 15, 16, 17]. Nonetheless, despite their effectiveness, these deep learning approaches may have poor data efficiency. They are solely data-driven and may not incorporate physical constraints, potentially resulting in spurious predictions. Moreover, these models cannot replace high-fidelity numerical simulations since they require a large quantity of high-quality snapshots for training. Physics-informed Neural Networks (PINNs) are another deep learning techniques that incorporate known physical laws or principles into their architecture, allowing them to learn from scarce data while satisfying the governing equations of the system. The basic idea behind PINNs and its variants is to compile the physics-based term to the loss function and take advantage of automatic differentiation to calculate derivative terms within the residual [18, 19, 20, 21]. By doing this, PINNs penalize solutions that violate physical laws and therefore improve the data efficiency. However, in the context of subsurface flow simulation, PINNs may face challenges when enforcing mass conservation in heterogenous porous media. This is due to the spatial variability of some reservoir properties, such as permeability and porosity. [22] proposed a network to estimate the non-uniform permeability field. Nevertheless, it is worth noting that additional errors may arise due to the scarcity of labeled data. [23, 24, 25] also employed PINNs for porous flow problems, however, their applications are limited to 1D problem or 2D cases with homogenous properties.

On the other hand, Physics-informed Convolutional Neural Networks (PICNNs), which leverage convolutional neural network to process spatial data, such as images and grids, can incorporate spatial heterogeneities naturally. Unlike PINN, PICNNs typically use convolutional filters derived from numerical techniques (e.g. finite difference, finite volume, finite element methods) to approximate the derivative terms in the governing equations. Regardless of the approach chosen, the primary objective is to enforce the principles of mass and energy conservation while maintaining the neural network's computational graph. [26] adopted a combination of elementary differential operators inspired by Fourier neural network for simulating multiphase flow in heterogenous porous media and suggest that it improves prediction accuracy than its PINN counterpart. [27] presents a framework called theory-guided convolutional neural network (TgCNN) for solving porous media flow problems. The TgCNN approach involves incorporating the residuals of discretized governing equations during the training of convolutional neural networks (CNNs) to improve their accuracy compared to regular CNN models. The trained TgCNN surrogates are further used for inverse modeling by combining them with the iterative ensemble smoother algorithm, and improved inversion efficiency is achieved with sufficient estimation accuracy. Reference [28] presents a deep convolutional model for simulating and predicting Darcy flows in heterogeneous reservoir models without requiring labeled data. It shows that their model can accurately simulate transient Darcy flows in homogeneous and heterogeneous reservoirs and can be trained as a surrogate to predict water breakthrough time if given sufficient amount of labeled data. However, these aforementioned approaches do not incorporate time-varying well controls to the model, which is a scenario more commonly encountered in practical reservoir management cases.

Inheriting the strengths of PICNNs and based on previous work [29], we presented a model that is able to emulate two-phase flow dynamics in heterogenous porous media with time-varying controls. Instead of coping with strong-form PDEs directly, our model is built upon discrete state-space equations. We use convolutional neural network so that heterogeneity can be incorporated on a pixel level, and we learn the network for each timestep in that the mapping from control to states is time variant. In contrast to previous approaches, our model can learn the discrete-time dynamics of state-space systems. The network takes initial reservoir state variables and a sequence of well controls as inputs and is trained by minimizing the residual of state-space equations of the system, which are the semi-discretized forms of the PDEs. The time derivative is calculated using backward Euler. The objective of this work is to predict reservoir flow quantities using image-to-image regression approach, with the input being the well controls (bottom hole pressure, injection rates) and output being state variables (pressures, saturations, etc.).

The rest of this paper is structured as follows. First, we introduce the governing equation for oil-water flow in porous media and its state-space representation. Second, we present a physics-informed convolutional neural network (PICNN) designed to learn the control-to-state mapping, along with our transfer learning approach. Next, we apply PICNN to simulate oil-water flow in porous media and compare the results with numerical simulators. Finally, we conclude this study and outline directions for future research.

## 2  A State-Space Perspective of Two-Phase Flow in Porous Media

### 2.1  Two-Phase (Oil-Water) Flow in Porous Media

In this study, two-phase (oil and water) porous flow is considered. Considering isothermal conditions, the resulting mass balance equations for each phase reads as follows,

$$\frac{\partial}{\partial t}(\frac{\rho_\alpha \phi S_\alpha}{B_\alpha}) + \nabla \cdot \left(\frac{\rho_\alpha \boldsymbol{v}_\alpha}{B_\alpha}\right) = \rho_\alpha q_\alpha$$

(1)

Where subscript $\alpha$ are used to identity phase of the fluid, with $o$ representing the oil phase and $w$ the water phase. $\rho_\alpha$ represents fluid density of phase $\alpha$, $\phi$ is the rock porosity, and $S_\alpha$ denotes the saturation of phase $\alpha$. In practice, the formulation is expressed in terms of surface volumes, so $B_\alpha$, referred as formation volume factor, is the quantity to convert in-situ volumes to surface volumes. $q_\alpha$ denotes the volumetric source/sink term at surface condition. Darcy's velocity for each phase can be expressed as:

$$\boldsymbol{v}_\alpha = -\frac{k_{r,\alpha}(S_\alpha)}{\mu_\alpha}\boldsymbol{K}(\nabla P_\alpha - \rho_\alpha g \nabla z)$$

(2)

Where $k_{r,\alpha}$ is relative permeabilities, which is a function of phase saturation $S_\alpha$. $\boldsymbol{K}$ denotes the permeability tensor, $\mu_\alpha$ is the fluid viscosity, $P_\alpha$ represents pressure for each phase, and gravity terms $\rho_\alpha g \nabla z$ are usually discarded for simplicity. Additionally, several auxiliary relationships that complete above equations include:

$$S_o + S_w = 1$$

(3)

$$P_o - P_w = P_c(S_w) \tag{4}$$

Oil-water capillary pressure $P_c(S_w)$ is negligible in reservoir scales, which implies that $P_o = P_w$. By introducing fluid and rock compressibility and pluging in Darcy's law, the equations can be rewritten as:

$$\frac{\rho_\alpha \phi}{B_\alpha}[S_\alpha(c_\alpha + c_r)\frac{\partial P_\alpha}{\partial t} + \frac{\partial S_\alpha}{\partial t}] + \nabla \cdot \left(-\frac{\rho_\alpha}{B_\alpha}\frac{k_{r,\alpha}(S_\alpha)}{\mu_\alpha}\boldsymbol{K}\nabla P_\alpha\right) = \rho_\alpha q_\alpha \tag{5}$$

Where $c_\alpha$ denotes the fluid compressibility of phase $\alpha$, and $c_r$ represents rock compressibility. For slightly compressible fluids, $\rho_\alpha$ fluid density is assumed to be constant and $c_\alpha$ to be finite. The governing equation is subject to different types of boundary and initial conditions, in our work, a no flow boundary condition (Neumann condition) is considered.

Integrating equation (5) over a representative elementary volume (REV) $V_i$ yields:

$$\underbrace{\int_{V_i} \frac{\phi}{B_\alpha}[S_\alpha(c_\alpha + c_r)\frac{\partial P_\alpha}{\partial t} + \frac{\partial S_\alpha}{\partial t}] \mathrm{dV}}_{\text{Accumulation term}} + \underbrace{\int_{V_i} \nabla \cdot \left(-\frac{1}{B_\alpha}\frac{k_{r,\alpha}(S_\alpha)}{\mu_\alpha}\boldsymbol{K}\nabla P_\alpha\right) \mathrm{dV}}_{\text{Flux term}} = \underbrace{\int_{V_i} q_\alpha \mathrm{dV}}_{\text{Source/Sink}} \tag{6}$$

The accumulation term can be rewritten as a control volume form: $V_i \frac{\phi}{B_\alpha}[S_\alpha(c_\alpha + c_r)\frac{\partial P_\alpha}{\partial t} + \frac{\partial S_\alpha}{\partial t}]$, the flux term can be expanded as an interface integral applying Gauss's theorem: $\int_{\partial V_i} -\frac{1}{B_\alpha}\frac{k_{r,\alpha}(S_\alpha)}{\mu_\alpha}\boldsymbol{K}\nabla P_\alpha \cdot \vec{n} \, \mathrm{dS} = -\sum_j T_{ij}(P_{j,\alpha} - P_{i,\alpha})$, subscript $j$ denotes the index of nearest neighboring grid cells connected to grid cell $i$. $T_{ij}$ is referred as transmissibility term between grid cell $i$ and $j$, which is expressed as:

$$T_{ij} = \frac{1}{B_\alpha}\frac{(k_{r,\alpha})_{ij}}{\mu_\alpha}K_{ij}\frac{A_{ij}}{d_{ij}} \tag{7}$$

Where $A_{ij}$ denotes the interface area between grid cell $i$ and $j$, $d_{ij}$ is the distance of two cell centers. The absolute permeability $K_{ij} = \frac{K_i K_j}{K_i + K_j}$ is calculated with harmonic average. To achieve accurate convective behavior, the relative permeability $(k_{r,\alpha})_{ij}$ is determined using upstream weighting:

$$(k_{r,\alpha})_{ij} = \begin{cases} (k_{r,\alpha})_i & \text{if } P_{i,\alpha} > P_{j,\alpha} \\ (k_{r,\alpha})_j & \text{else} \end{cases} \tag{8}$$

$\int_{V_i} q_\alpha \, \mathrm{dV} = Q_{i,\alpha}$ is the source/sink term. In the field of reservoir simulation, either flow rate or bottom hole pressure (BHP) can be prescribed. Flow rate may be incorporated directly, while BHP is prescribed indirectly using a Peaceman well model [30] at the well block $i$, which is expressed as:

$$q_{i,\alpha} = \frac{k_{r,\alpha}(S_\alpha)}{\mu_\alpha} WI_i \times (P_{i,\alpha} - P_{wf})$$

(9)

$P_i$ is the pressure at the well block and $P_{wf}$ the prescribed BHP, $WI_i$ is the well index at the well block $i$ and for an isotropic reservoir where $K = K_x = K_y$, the well index reads

$$WI_i = \frac{2\pi\alpha K_i h}{\ln(\frac{r_e}{r_w} + s)}$$

(10)

Where $\alpha$ is a unit conversion factor, $r_e = 0.14(\Delta x^2 + \Delta y^2)^{1/2}$ is the effective radius of the well block, $r_w$ represents the wellbore radius and $s$ denotes the skin factor of the wellbore.

Assembling all grid blocks at the computational domain yields the following matrix form:

$$\begin{bmatrix} A_{op} & A_{os} \\ A_{wp} & A_{ws} \end{bmatrix} \begin{bmatrix} \dot{P} \\ \dot{S}_w \end{bmatrix} + \begin{bmatrix} T_{op} & 0 \\ T_{wp} & 0 \end{bmatrix} \begin{bmatrix} P \\ S_w \end{bmatrix} = \begin{bmatrix} Q_o \\ Q_w \end{bmatrix}$$

(11)

It is worth noting that Equation (11) is nonlinear in that coefficients of all sub-matrices are functions of $S_w$ or $P$, e.g. relative permeabilities in $T_{op}$ and $T_{wp}$ are dependent on saturation, and fluid viscosities and formation volume factors are a function of fluid pressure, etc. In a typical numerical simulator, equation 11 is rearranged to an algebraic form can is solved iteratively until desired accuracy is reached.

## 2.2 State Space Representation

Understanding reservoir flow equations from a system and control perspective is crucial in designing and implementing effective well control strategies that improve production efficiency and reduce operational costs in subsurface reservoirs [31]. Moreover, it can also guide us to design effective neural network architectures. Committing to the systematic approach, a state-space representation of equation 11 can be written as follows:

$$\dot{x}(t) = A_c(x,t)x(t) + B_c(x,t)u(t)$$
$$y(t) = C_c(x,t)x(t) + D_c(x,t)u(t)$$

(12)

Where $x = \begin{bmatrix} P \\ S_w \end{bmatrix}$ represents the state, $u = \begin{bmatrix} P_{wf} \\ q_{inj} \end{bmatrix}$ is the system inputs/controls, $\dot{x}$ denotes the time derivative of $x$, $A_c$ is referred as state matrix and $B_c$ is called system input matrix. $A_c$ and $B_c$ are a function of state variable $x$.

The practical computational implementation when dealing with time-varying inputs/controls $u$ typically requires temporal discretization of continuous state-space equation 12. Applying implicit Euler scheme, we obtain

$$x_{k+1} = x_k + \Delta t(A_c x_{k+1} + B_c u_{k+1})$$

$$y_{k+1} = C_c x_{k+1} + D_c u_{k+1}$$

(13)

By rearranging 13, it leads to discrete-time form

$$x_{k+1} = A_d x_k + B_d u_{k+1}$$

(14)

Or in a more general form

$$x_{k+1} = f_{k+1}(x_{k+1}, x_k, u_{k+1})$$

(15)

Here one set of signals $U = \{u_1, u_2, \ldots, u_t\}$ with given initial state $x_0$ will lead to a unique set of states $X = \{x_1, x_2, \ldots, x_t\}$, which is known as one trajectory. Some efforts have been made to use limited state trajectories to discover unknown dynamical system mechanics [32, 33], however, this still relies on sufficient labelled data for training. The objective of this work is to build a signal-to-state ($u \sim x$) mapping using neural networks without any labelled data. Since $u$ is time-varying, and the mapping from $u_k$ to $x_k$ is changing over time, traditional neural networks that sharing weights across all timesteps (RNNs, LSTM, etc.) may not be capable of modeling the time-varying dynamics. Therefore, we propose a network that leverages transfer learning and PICNN to model the complicated relationships in an efficient and label-free way.

## 3    Physics-informed Convolutional Neural Network (PICNN)

### 3.1    PICNN framework and architecture

U-Net is a specialized deep learning architecture designed primarily for segmentation tasks, especially in the field of biomedical image analysis [34, 35]. It has an encoder-decoder framework consisting of a contracting path (encoder) and an expansive path (decoder). The encoder is made of a series of convolutional and polling layers. This path progressively reduces the spatial dimensions of the input image while increasing the number of feature maps. The decoder part involves a symmetrical structure to the contracting path. U-Net also introduces skip connections, which are direct connections between the contracting and expansive paths at corresponding levels. These connections allow the network to retain fine-grained information from the contracting path. U-Net has demonstrated remarkable success in various computer vision and engineering applications, including medical image segmentation, data-driven surrogate modeling, etc. [36, 37].

The architecture of the convolutional neural network used in our work for each timestep is depicted in Figure 1. The architecture is based on U-Net and is meticulously modified for our specific task. As shown in Figure 1, we employ a parallel U-net structure to predict pressure and saturation separately to acknowledge the significant difference in magnitude (at least in terms of field units) between these two quantities. The architecture for both the pressure and saturation path are identical except the final scaling layers. These scaling layers are employed to restrict the space of acceptable solutions given prior physical knowledge, e.g. the water saturation should fall within the range [$S_{wc}$, 1-$S_{or}$], and the pressure should typically not be lower than $P_{wf}$. Mathematically, the final saturation output by network is expressed as

$$S_w^{NN} = S_{wc} + (1 - S_{or} - S_{wc}) * x_{s_w}$$

Where $x_{s_w} \in [0, 1]$ represents the output from previous sigmoid layer. A similar formula applies to the pressure

$$P^{NN} = P_{min} + (P_{max} - P_{min}) * x_P$$

Where $P_{min}$ is some value smaller than $P_{wf}$, and $P_{max}$ is the maximum value that the system can reach. The water saturation must be within a strict range because values outside this range would cause the network to stop learning. This behavior can be understood by referring to the relative permeability plot in Figure 4. This plot shows that the rate of change of relative permeabilities concerning water saturation becomes zero when the saturation is either less than 0.2 or greater than 0.8, resulting in vanishing gradients. In contrast, the pressure constraint is less stringent, as the quantities change gradually with pressure. It is evident that the inclusion of the scaling layer contributes to the network's faster convergence because it further restricts the space of acceptable solutions.

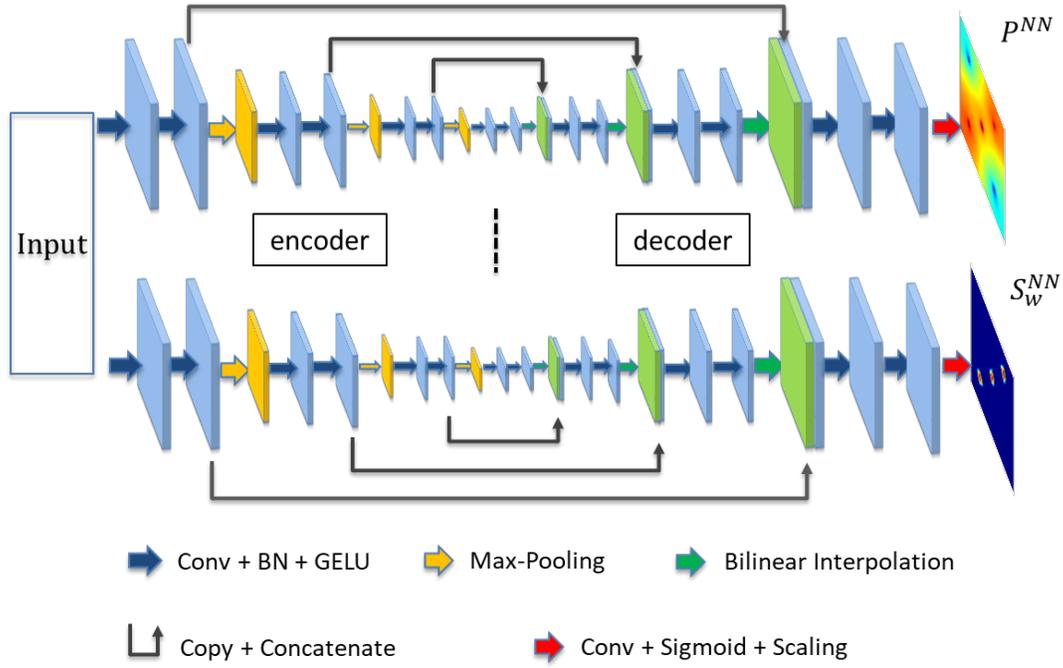

**Figure. 1.** Proposed network architecture for each timestep. The input consists of a two-channel map containing information about BHP (bottle hole pressure) and injection rate. The blue arrows correspond to convolutional operation (kernel size $3 \times 3$, padding size 1) followed by batch normalization and GELU activation. Each yellow arrow represents a $2 \times 2$ max pooling with a stride of 2 for downsampling. During downsampling, the number of feature channels is doubled. Conversely, in the expansive path, the green arrows indicate upsampling operations that increase the dimension by 2 while reduces the number of feature channels by half. Both downsampling and upsampling are performed 3 times in this network. Each black elbow arrow denotes skip connections, which establish direct connections between the contracting and expansive paths at corresponding levels. Finally, at the last layer (red arrow) a $1 \times 1$ convolutional followed by a sigmoid and scaling layer is used.

## 3.2 Transfer learning-based PICNN

Since the relationship between control and state varies with time, employing neural networks with sharing weights across all timesteps may limit their ability to capture the time-changing dynamics. Consequently, a deep neural network that mimics the discretized state-space equation is proposed to learn the control to state mapping time-step wisely. The temporal domain is subdivided into equidistant time grids, and a sequence of convolutional neural networks is utilized, with each CNN dedicated to a specific time instance. The CNN outputs are used to update state-dependent fluid and rock properties, which are subsequently used to construct the global accumulation and transmissibility matrices and formulate the loss for that timestep. The training of the PICNN for a given time step is completed before its outputs are advanced to the subsequent time step. This sequential progression is essential as past states are necessary for calculating the loss at the current time step, as illustrated in equations 14 and 15. Additionally, rather than training models from scratch, we employ transfer learning for subsequent time steps once the network from the previous time step is fully trained. The comprehensive network architecture is visualized in Figure 2.

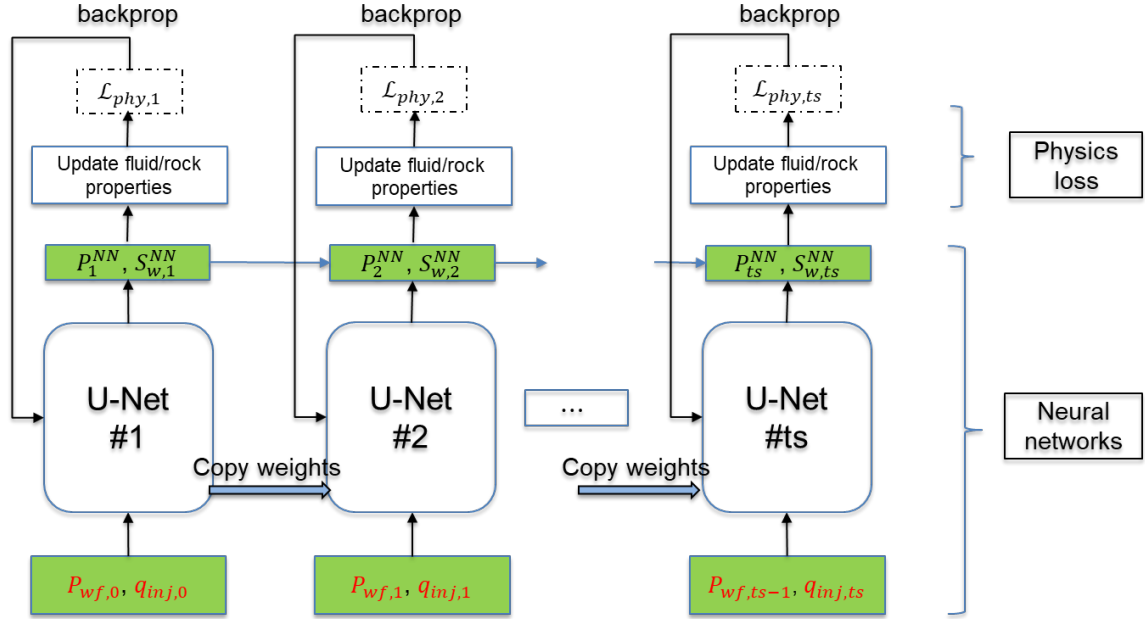

**Figure. 2.** Transfer learning-based PICNN for all timesteps. The strcuture includes a collective of $ts$ U-Nets presented in Figure 1, where each U-Net is specifically assigned to one timestep. Each network is trained solely by minimizing the physics loss for their respective timestep. Upon completion of training for one network, both the output of the network and its weights are proceeded to the subsequent timestep.

## 3.3 Physics-informed Loss and Training

The learning process is unsupervised since no labeled data is required. The network is trained by minimizing the physics loss at each timestep, forcing the admissible solution space comply with the domain knowledge. The physics loss at each timestep can be minimized in two ways:

$$\mathcal{L}_{phy,k} = \mathcal{L}_s(r_k, 0) \quad \text{or} \quad \mathcal{L}_{phy,k} = \mathcal{L}_s(r_{o,k}, 0) + \mathcal{L}_s(r_{w,k}, 0)$$

The latter one is used in our study from a memory efficiency standpoint. The total residual $r_k$ and residuals of each phase $r_{o,k}$, $r_{w,k}$ are defined as

$$r_k = \begin{bmatrix} r_{o,k} \\ r_{w,k} \end{bmatrix} = \begin{bmatrix} A_{op} & A_{os} \\ A_{wp} & A_{ws} \end{bmatrix} \begin{bmatrix} \dot{P}^{NN} \\ \dot{S}_w^{NN} \end{bmatrix}_k + \begin{bmatrix} T_{op} & 0 \\ T_{wp} & 0 \end{bmatrix} \begin{bmatrix} P_k^{NN} \\ S_{w,k}^{NN} \end{bmatrix} - \begin{bmatrix} Q_o \\ Q_w \end{bmatrix}$$

$\mathcal{L}_s$ represents a smooth $L_1$ loss function operator. $P_k^{NN}$, $S_{w,k}^{NN}$ is the network output at timestep $k$. The $\mathcal{L}_s$ has the following formulation:

$$\mathcal{L}_s(y_{pred}, y_{true}) = \begin{cases} \dfrac{0.5(y_{pred} - y_{true})^2}{\beta}, & if |y_{pred} - y_{true}| < \beta \\ |y_{pred} - y_{true}| - 0.5 \times \beta, & otherwise \end{cases}$$

Where $\beta$ is a hyperparameter, which is a fixed threshold that divides $\mathcal{L}_s$ loss into $\mathcal{L}_1$ and $\mathcal{L}_2$ regions. In our work, $\beta$ is set to be 10. The $\mathcal{L}_s$ function behaves like the $L_2$ loss function when the absolute difference between the predicted and true values is small, and like the $L_1$ loss function when the absolute difference is large. The Smooth L1 loss offers more stable training because it has bounded and linear behavior for large errors [38].

Adam optimizer is employed to train the network. The initial learning rate $\eta$ is set to 0.01 and is decayed step-wise with a parameter of 0.995 for every 100 epochs. Network training process is depicted in figure 3. All experiments were conducted using a NVIDIA A100 GPU provided by Texas A&M High Performance Research Computing. We have summarized all training steps in Algorithm 1. This algorithm progressively trains a neural network to simulate and predict the behavior of porous flow system over multiple timesteps while ensuring that the predictions adhere to the underlying physical laws. It also demonstrates how neural networks can be used to capture complex dynamics in physical systems and how transfer learning is applied from one timestep to the next to expedite training. The training ceases if either the desired training accuracy $\sigma$ is met or the maximum training epoch is reached.

**ALGORITHM 1: TRAINING PHYSICS-INFORMED CONVOLUTIONAL NEURAL NETWORK**

**Input:** total simulation time $T$; timestep size $\Delta t$; total timesteps $ts = T/\Delta t$; Neural network model $NN_k(\ldots, \theta_k)$ for each timestep $k = 1, 2, \ldots, ts$; initial condition $x_0 = \{P_0, S_{w,0}\}$; a sequence of well controls $U = \{[P_{wf,0}, q_{inj,0}], [P_{wf,1}, q_{inj,1}], \ldots, [P_{wf,ts}, q_{inj,ts}]\}$; learning rate $\eta$; maximum training epochs $N_{epoch}$ and desired training accuracy $\sigma$

**for** k=1 **to** $ts$ **do**
    **if** k > 1 **do**
        $\theta_k \leftarrow \theta_{k-1}$     # initialize network weights with weights from previous timestep
        $P_{k-1}^{NN}, S_{w,k-1}^{NN} \leftarrow NN_{k-1}(U_{k-1}, \theta_{k-1})$
    **else**
        $\theta_k \leftarrow$ Kaiming initialization
        $P_{k-1}^{NN}, S_{w,k-1}^{NN} \leftarrow P_0, S_{w,0}$
    **end**
    **while** loss $> \sigma$ **and** epoch $< N_{epoch}$ **do**
        $P_k^{NN}, S_{w,k}^{NN} \leftarrow NN_k(U_k, \theta_k)$     # forward pass of the model, $x_k^{NN} = [P_k^{NN}, S_{w,k}^{NN}]$
        Update $\mu_o, \mu_w, k_{ro}, k_{rw}$ based on $P_k^{NN}, S_{w,k}^{NN}$
        $\mathcal{L}_{phy,k} = \mathcal{L}_s(x_k^{NN} - f_k(x_k^{NN}, x_{k-1}^{NN}, U_k), 0)$
        $\nabla \theta_k \leftarrow Backprop(\mathcal{L}_{phy,k})$
        $\theta_k \leftarrow optimizer.step()$     # update network weights using predefined optimizer
    **end**
**end**
**Output:** optimized model parameters $\{\theta_k\}$ for $k = 1, 2, \ldots, ts$

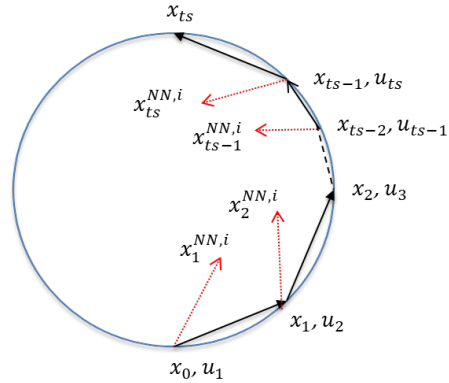

**Figure 3**: Learning process within the state-space manifold. The black lines represent a single state trajectory in response to a time-series of controls. The red lines depict the network's learning process for each respective timestep, with the superscript $i$ indicating the iteration step during training. Training ceases either when the desired accuracy is achieved or when the maximum training epoch is reached.

## 4    Case studies

In this section, we describe the model configuration for simulating oil-water flow and validate the PICNN-based simulation results against in-house simulators. To improve the accuracy of predictions for localized quantities like well block pressure and well rates, we apply additional regularization to the original PICNN based on production data. We also evaluate the computational efficiency of PICNN by varying the number of computational grid blocks. Finally, we assess the accuracy of the trained PICNN proxy for predicting unseen trajectories, presenting the error results for 10 different time-series controls.

### 4.1 Oil-water flow in heterogenous reservoir – small example

The reservoir model consists of 64 × 64 grid blocks, each of which has dimension of 20m×20m×20m in x, y and z directions. The permeability field is primarily from the SPE10 dataset and is plotted in Figure 5. Reservoir initial pressure is 3000 psia, and the initial irreducible water saturation is set to be $S_{wc} = 0.2$. The reservoir is operated by three injection wells (source) and two production wells (sink), with their well locations depicted in Figure 5. The injection wells inject water into the system to supplement the reservoir pressure, while the producers extract oil from the reservoir. A no-flow boundary condition is considered for all experiments. Other important physical properties used in the simulation are listed in Table 1.

**Table 1:** Physical properties

| Rock properties | | Fluid properties | | Wellbore properties | |
|---|---|---|---|---|---|
| porosity | 0.20 | Oil Viscosity, cp | 1.13 | Well radius, ft | 0.30 |
| Initial pressure, psia | 3000 | Oil compressibility, $psia^{-1}$ | $1.0 \times 10^5$ | skin factor | 0.0 |
| Rock compressibility, $psia^{-1}$ | $3.0 \times 10^6$ | Water compressibility, $psia^{-1}$ | $3.0 \times 10^6$ | | |

Corey model is used to calculate saturation-dependent relative permeability, which is expressed as:

$$k_{rw} = k_{rw}^0 S^{n_w}$$

$$k_{ro} = k_{ro}^0 (1-S)^{n_o}$$

Where $k_{rw}^0$ and $k_{ro}^0$ are end-point relative permeabilities, $S = \frac{S_w - S_{wc}}{1 - S_{or} - S_{wc}}$, $S_{wc}$ is connate water saturation and $S_{or}$ is the residual oil saturation. $n_w$ and $n_o$ are Corey exponents. In our case, $S_{wc} = 0.2$, $S_{or} = 0.2$, $n_w = 2$, $n_o = 3$, $k_{rw}^0 = 0.60$ and $k_{ro}^0 = 0.90$.

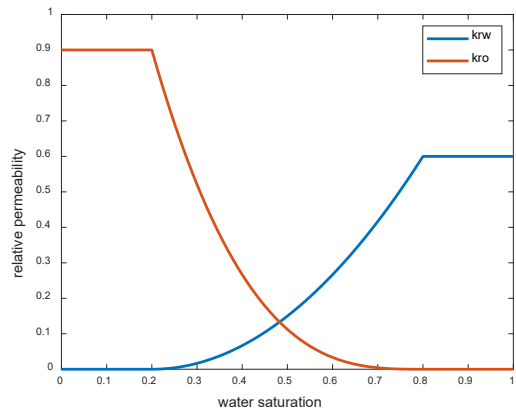

**Figure 4:** Relative permeability for oil phase (orange) and water phase (blue).

Total simulation time is 100 days with equidistant time step size 2 days. The injection wells are controlled by specifying the stepwise injection rates while the production wells are controlled by prescribing stepwise BHPs (figure 5). These control parameters for both injection and production wells undergo stepwise changes every 50 days. The high-fidelity simulations are performed using a locally hosted CPU with an Intel(R) Xeon(R) W-2223 CPU @3.60GHz.

The PICNN is fully trained in a way that either the training loss is smaller than predefined training accuracy $\sigma$ or the maximum training epoch is reached. After the offline training stage, reservoir states can be inferred sequentially. We then compare the predicted pressure and saturation using trained PICNN against the reference ones from numerical simulator at 20, 40 and 100 days, as shown in Figure 6. The relative error map is defined as $\frac{X_{pred}-X_{ref}}{X_{ref}}$ to compare the accuracy.

Figure 6 presents the results of the PICNN model's predictions for pressure and saturation snapshots in the above heterogeneous reservoir under varying well control scenarios, along with the corresponding reference pressure and saturation profiles obtained from in-house reservoir simulator and relative error maps. The results at selected days demonstrate that the pressure and saturation maps predicted by the PICNN model are in decent agreement with the reference ones from the numerical simulator. For the pressure maps, the relative error maps suggest slightly higher errors at extraction regions—around 2.5% at the most. For water saturations, relative error maps show that most of the error is located near the injection wells, particularly at the water front region. For other regions, both pressure and saturation almost show perfect match, with relative error is near 0.

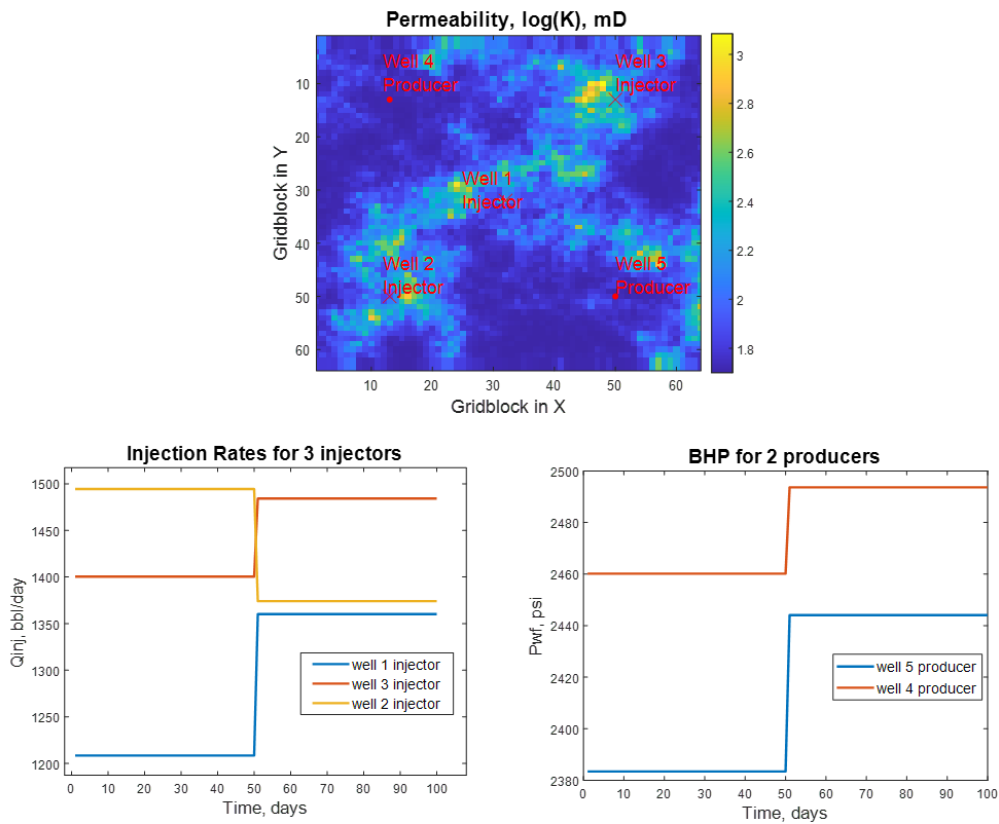

**Figure 5:** Permeability map of reservoir and well locations (top);  Injection rates of injection wells (bottom left);    BHP schedule of production wells (bottom right)

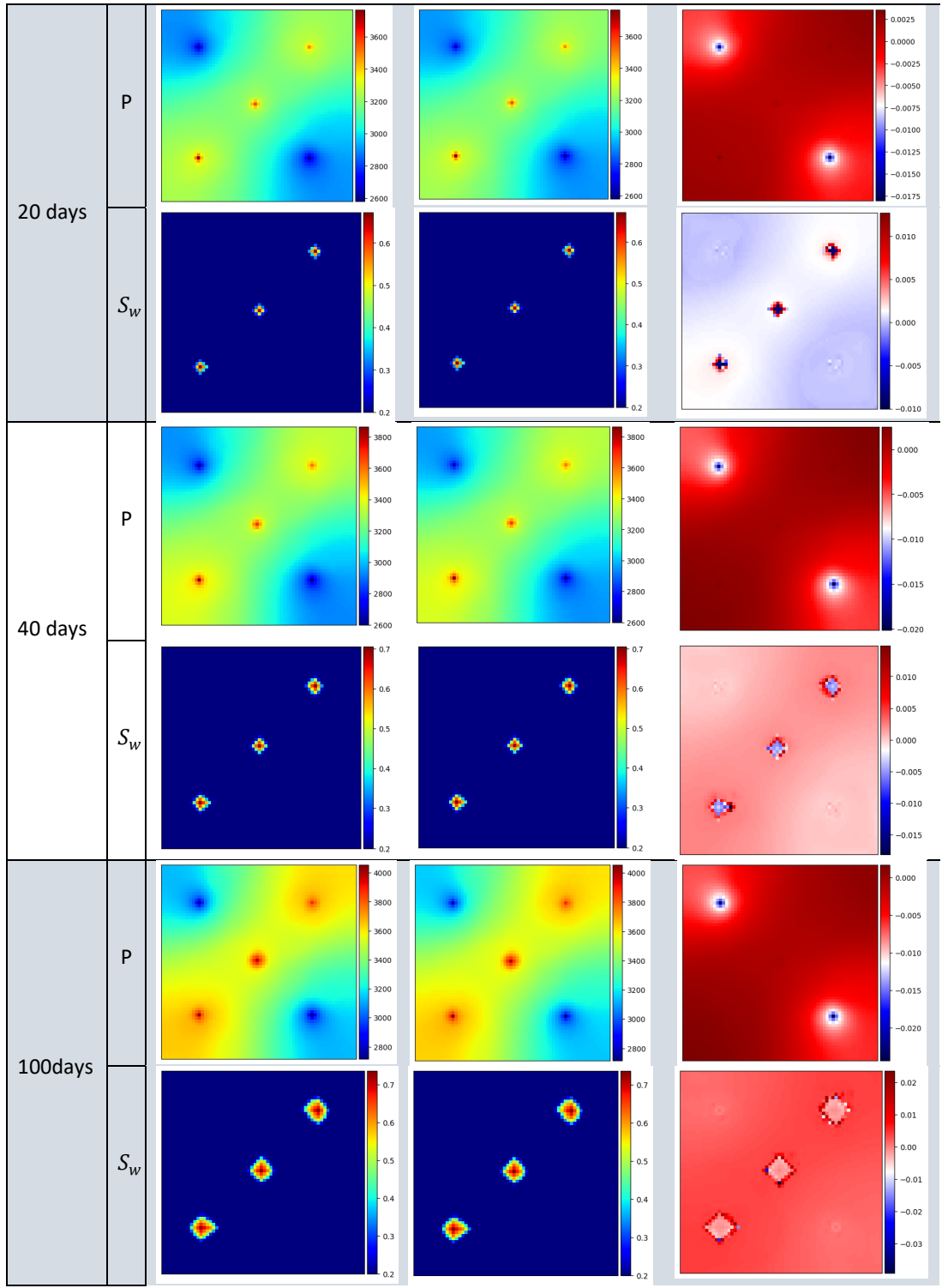

**Figure 6:** Predicted pressure/saturation with PICNN when $\sigma = 0.002$ (left column), Reference pressure/saturation from numerical simulator (middle column), and relative error (right column) at different time T=20, 40, 100 days.

In order to evaluate the impact of various training accuracies on computational efficiency, we conducted a comparison of the training times associated with three different training accuracies denoted as $\sigma$. The resulting histogram of training times for these different $\sigma$ values is illustrated in Figure 7 (left). Specifically, for $\sigma = 0.002$, the training process required 7590 seconds, for $\sigma = 0.005$, it took 3345 seconds, and for $\sigma = 0.05$, the training duration was 2010 seconds. To quantify the influence of different training accuracies on prediction accuracy, we utilized the Mean Absolute Percentage Error (MAPE) metric, which measures the disparity between predicted states and reference states. The MAPE is defined as:

$$MAPE(y, \hat{y}) = \frac{1}{N} \sum_{i=1}^{N} \frac{|y_i - \hat{y}_i|}{\max(|y_i|)}$$

Where $y$ is the pressure or saturation and $\hat{y}$ is corresponding reference pressure and saturation, $N$ denotes the dimensionality of $y$. This MAPE evaluation over time is depicted in Figure 7 (right).

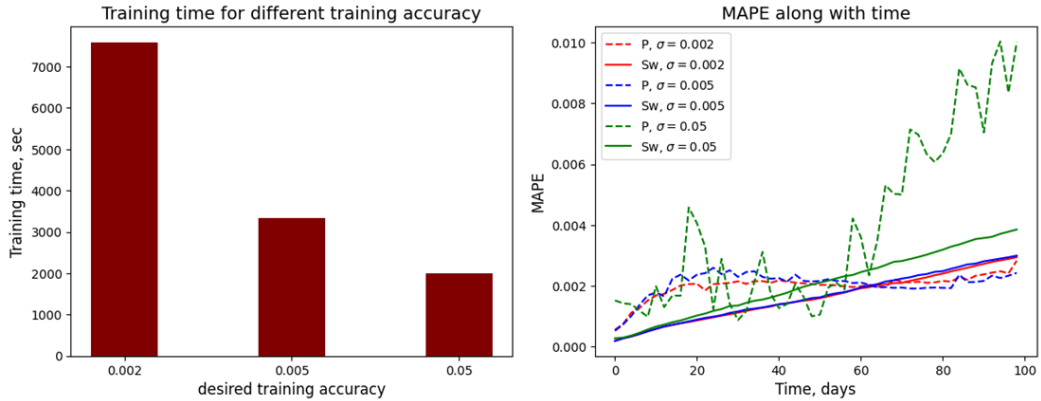

**Figure 7:** Total training time for different training accuracy $\sigma$ (left); Mean absolute percentage error along with time (right).

To sum up, lower training accuracy results in faster training of the proposed PICNN model. When comparing $\sigma = 0.002$ and $\sigma = 0.005$, there isn't a significant difference observed in the temporal MAPE. However, for $\sigma = 0.05$, the MAPE for pressure closely aligns with the values obtained for smaller $\sigma$, while a slightly higher MAPE is observed for the saturation profile. Nonetheless, this performance is considered satisfactory given the inherent strong nonlinearity in saturation behavior. It's important to note that the choice of training accuracy depends on the specific loss function used and should be determined through experimental evaluation.

### 4.2   PICNN regularized on production data

In section 4.1 we showed PICNN has good performance in simulating and predicting global quantities such as the entire pressure and saturation maps. In practical application, there is often a greater interest in local quantities such as well block pressure (WBP), well production rates, etc. Figure 6 illustrates that the predicted WBPs by PICNN are consistently lower than the reference values. In fact, we observe a systematic bias in the predicted well block pressures (WBPs) across all timesteps with the original PICNN, as shown in Figure 8. To enhance the accuracy of predictions for quantities at well locations, we introduce a regularization technique for PICNN. This regularization involves incorporating production data by adding a data loss term, which is formulated as follows:

$$\mathcal{L}_{total,k} = \alpha \mathcal{L}_{phy,k} + \beta \mathcal{L}_{data,k}$$

Where $\mathcal{L}_{data,k} = \frac{1}{N_{prod}} \sum_{i=1}^{N_{prod}} L_1(WBP_{i,k}^{NN} - WBP_{i,k})$. α, β are trainable weights for physical loss and data loss. In this work, the initial α and β is set to be 1 and 0.01, respectively. The PICNN is now trained with $\mathcal{L}_{total,k}$ instead of only physical loss.

Figure 8 displays the well block pressures (WBPs) and production rates for two production wells, both before and after the application of regularization. When utilizing the original PICNN, both WBPs and production rates deviate from the reference curves. Predicted WBPs are consistently 40 to 60 psia lower than the reference WBPs, and oil rates are underestimated by 250 to 350 STB/day compared to the reference rates. Upon the introduction of the production data loss, a noticeable improvement is observed. Both WBPs and oil rates now align perfectly with the true values.

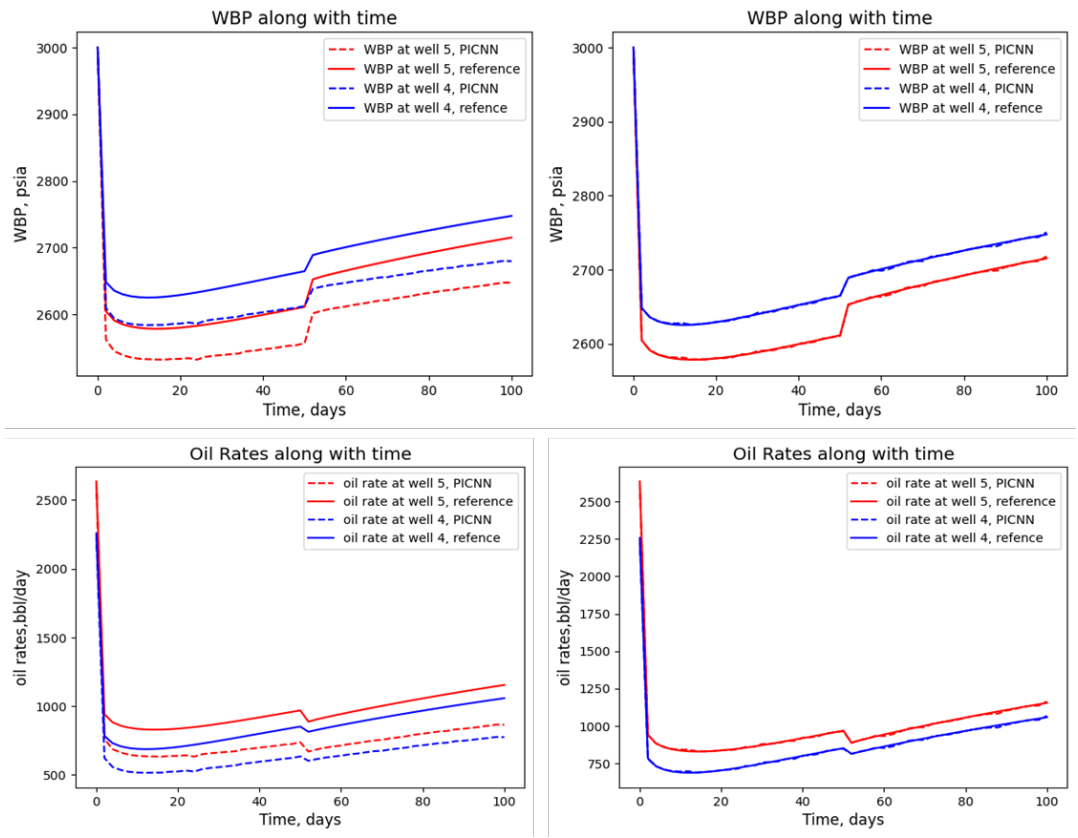

**Figure 8:** WBPs from original PICNN and reference (top left); WBPs from PICNN regularized on production data and reference (top right). Oil rates from original PICNN and reference (bottom left); Oil rates from PICNN regularized on production data and reference (bottom right).

## 4.3 PICNN for different reservoir model dimensionality

In this subsection, we explore the computational aspects of the proposed PICNN by varying the number of grid cells in the reservoir model. In traditional numerical simulator, it is intuitive that the computational time is directly proportional to the number of degree of freedoms (DoFs). This relationship arises because a higher number of DoFs involves larger matrix evaluations and inversions within the algebraic equations. On the other hand, the computational cost of a PICNN is influenced by a combination of factors related to the network architecture, optimization technique, physics-informed constraints, hardware efficiency and so on. In this specific experiment, we solely change the dimensions of the input data (the grid numbers), keeping the network architecture (Figure 1) and other parameters of the network, such as the number of layers, the size of convolutional kernels and number of filters in each layer, fixed. Despite the change in reservoir model dimensionality, we observe that the training time for the PICNN remains unaffected. This is attributed to the fact that the computational complexity of the PICNN primarily depends on the number of trainable parameters in the network, and these parameters remain constant regardless of the reservoir's dimensionality. It's noteworthy that while larger input data sizes might necessitate deeper layers and increased filter numbers per layer in some scenarios, our experiments demonstrate that, in our specific cases, sophisticated CNN architectures like U-Net proved adequate for capturing the intricate relationships between input and output, illustrating the robustness and efficiency of our approach in handling varying input data sizes.

Figure 9 showcases the results of the PICNN model's predictions for pressure and saturation snapshots with different number of grid blocks at 100 days. The corresponding reference pressure, saturation, and relative error maps are also included in this experiment. In this experiment, the training accuracy $\sigma$ is set to be 0.05. Since our primary interest is the computational efficiency, for case 2 and case 3, we consider homogenous properties. The results clearly display that the pressure and saturation maps predicted by the PICNN model align well with the reference data obtained from the numerical simulator. In the pressure maps, the largest relative errors are situated near the source/sink regions, typically within a range of ±2.5%. On the other hand, for water saturations, the relative error maps indicate slightly larger errors, primarily near the water front. However, these errors are only confined to a few pixels, reaffirming the overall accuracy and reliability of the predictions.

In Table 2, we conducted a comparative analysis of two key computational metrics between PICNN and an in-house numerical simulator. In the context of a two-phase oil-water flow simulator with a total grid block count of $N_t$, the degree of freedom is typically equal to $2 \times N_t$. As the model's dimensionality increases, the simulation time experiences a significant rise. As shown, the total simulation time escalates from 20.3 seconds to 141.0 seconds. It's worth noting that the actual simulation time can vary depending on the specific simulator used in the study. Commercial simulators, for instance, may exhibit faster performance due to their advanced numerical solvers, preconditioners, and improved workload management techniques. However, it's important to emphasize that the primary aim of this study is not to compete with commercial simulators but rather to illustrate the potential of PICNN in addressing large-scale multi-physics problems. Table 3 also highlights that the number of trainable parameters in the PICNN remains constant, as the same CNN model is utilized across all three cases. Interestingly, as we increase the model's dimensionality, both training time and inference time do not experience as significant an increase as observed in the simulator. Across all three cases, PICNN achieved speedups of 1.75, 4.93, and 19.7 times during the online-stage inference respectively, showcasing its efficiency in making predictions for larger-scale systems.

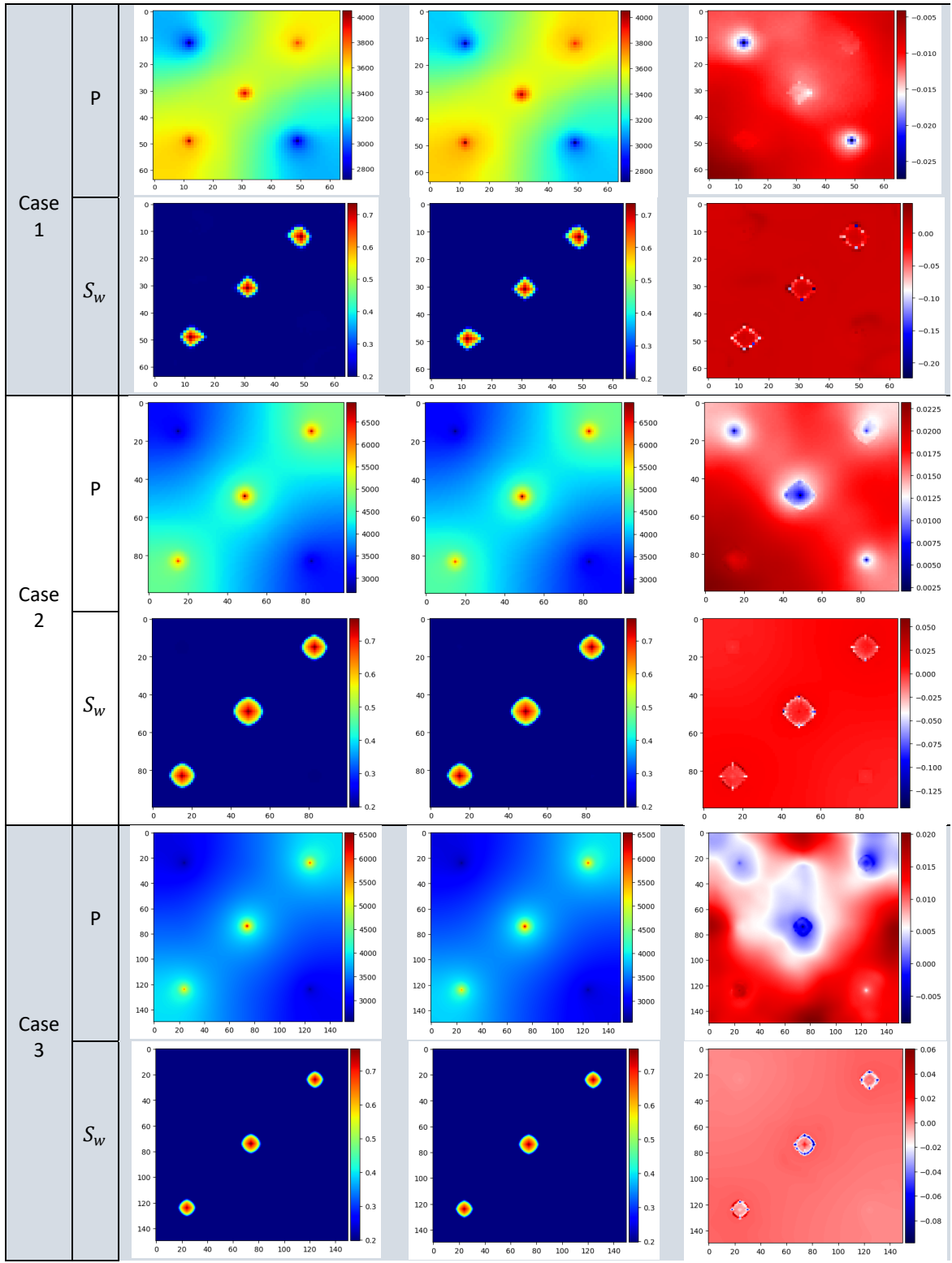

**Figure 9:** Predicted pressure/saturation with PICNN (left column), Reference pressure/saturation from numerical simulator (middle column), and relative error (right column) for cases with different number of grid blocks (Case 1: 64 by 64, Case 2: 100 by 100, Case 3: 150 by 150).

**Table 2:** Computational metrics of both PICNN and numerical simulator for different number of grid cells.

| Case | # of grids | PICNN ($\sigma = 0.05$) | | | Numerical Simulator | |
|---|---|---|---|---|---|---|
| | | # of trainable parameters | Training time, sec | Inference time, sec | Degree of freedoms (DoFs) | Simulation time, sec |
| 1 | 64×64 | 10862850 | 2009.9 | 11.57 | 8192 | 20.3 |
| 2 | 100×100 | 10862850 | 2810.2 | 11.62 | 20000 | 57.3 |
| 3* | 150×150 | 10862850 | 1058.1 | 7.17 | 45000 | 141.0 |
| * for case 3, the step size is set to 3 days. | | | | | | |

### 4.4 PICNN to predict unseen data

One advantage of establishing a control-to-state regression is the potential to predict states for different control scenarios. In other words, we could possibly deploy the trained model to predict unknown states for different control schedules. However, we have observed that this approach is effective only when the controls do not vary significantly. When there is a substantial variation in either the magnitude or frequency of the time-series controls, it can lead to inaccurate state predictions. To test this, we varied the injection rates between 1000 and 1500 STB/day and the bottom hole pressure (BHP) between 2300 and 2500 psia. We generated 10 different sets of testing controls within these ranges. Initially, these controls were altered every 50 days, as described in Section 4.1. Then, we significantly increased the alteration frequency by changing it to every 10 days.

Figure 10 illustrates the temporal evolution of the mean absolute percentage error (MAPE) in cases where control alterations occur every 50 days, just like the training case. In the left subplot, it depicts the MAPE progression for predicted pressure, with the baseline case represented by the black curve and 10 alternative time-series controls indicated by the dashed red curves. Meanwhile, in the right subplot, the MAPE evolution for predicted water saturation is displayed, with the black curve denoting the baseline case and the dashed blue curves signifying 10 distinct time-series controls. Moving on to Figure 11, it presents the temporal mean absolute percentage error (MAPE) for scenarios where control changes occur every 10 days. In the left subplot, the MAPE dynamics for predicted pressure are showcased, featuring the baseline case as the black curve and 10 alternative time-series controls represented by dashed red curves. In the corresponding right subplot, the MAPE evolution for predicted water saturation is exhibited, with the black curve corresponding to the baseline case and the dashed blue curves symbolizing 10 different time-series controls. From Figure 10 and 11 we observe that pressure MAPEs for these 10 cases are approximately an order of magnitude higher compared to the baseline case, while the phase saturation MAPEs are about twice as high as those in baseline model. Increasing frequency of well controls change would leads to more fluctuating and larger errors for both pressure and saturations. These errors can be elucidated by the state-space equations, which indicate that the mapping from well controls to reservoir states is not only time-dependent but also unique, contributing to these observed variations in error magnitudes.

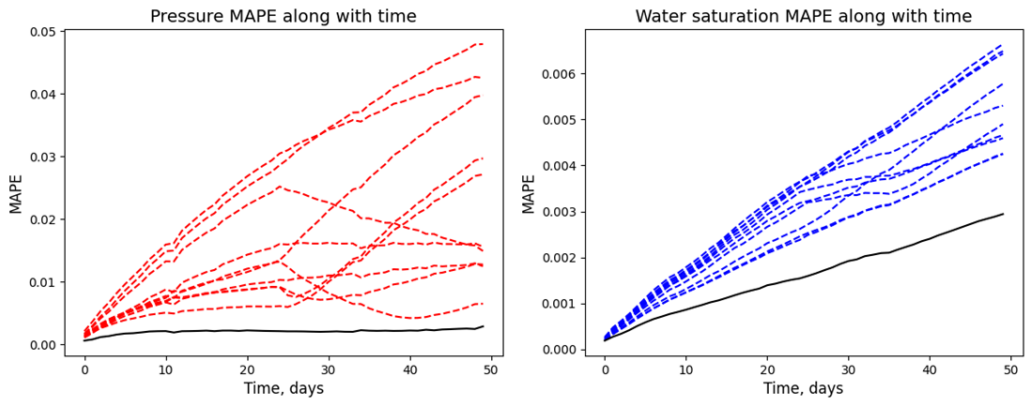

**Figure 10:** Temporal Mean Absolute Percentage Error (MAPE) of predicted pressure and water saturation when controls vary every 50 days.

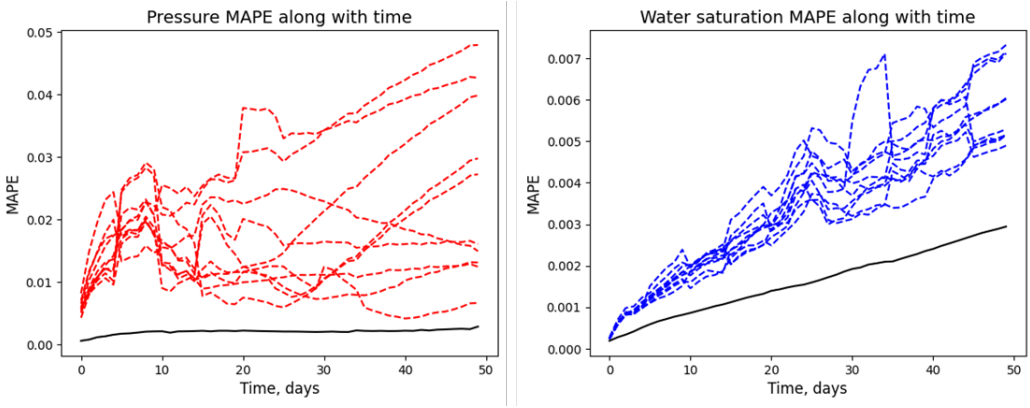

**Figure 11:** Temporal Mean Absolute Percentage Error (MAPE) for predicted pressure and water saturation when controls alter every 10 days.

## 5	Conclusions

In this work, a physics informed convolutional neural network (PICNN) is proposed to simulate and predict reservoir dynamics with time-varying well controls. The network is trained progressively to simulate two phase flow in porous media at each timestep without labeled data. Unlike many existing deep learning models that focus on mapping parameters to states, our model establishes a control-to-state mapping, making it particularly suitable for reservoir dynamical controls.

Our findings demonstrate that the proposed PICNN model performs well in predicting reservoir states. It achieves remarkable accuracy, with relative errors of less than 1% in regions far from source/sink areas for both pressure and water saturation predictions, and approximately 3% in areas near the wells. By incorporating production data into the training process, we significantly improve the estimation of local quantities like well block pressures (WBPs) and well rates. Furthermore, we explore various reservoir simulation scenarios with different grid block configurations and compare the computational efficiency and accuracy of PICNN against traditional numerical techniques. Notably, the PICNN proxy exhibits impressive speedups during online inference, showcasing its potential for efficiently handling large-scale reservoir models. Importantly, the computational costs associated with training the PICNN are not influenced by the dimensionality of the reservoir model, highlighting its suitability for tackling complex reservoir simulations efficiently. Lastly, we tested the proxy model with 10 different time-series controls and provided the error plots. We showed that the proposed model may not be robust and reliable if the controls vary significantly either in magnitude or frequency.

Our forthcoming research endeavors will prioritize the development of more robust and precise surrogate models, taking a dynamical system control perspective into account. Additionally, we intend to expand our methodology to encompass the simulation of more intricate 3D reservoir models characterized by a higher number of grid blocks. It's important to acknowledge that the proposed PICNN approach is currently applicable to Cartesian grids, which facilitate convolutional operations. In the future, we also plan to explore the potential of graph neural networks to address scenarios involving unstructured meshes.

**ACKNOWLEDGEMENTS**

Portions of this work were conducted with the advanced computing resources provided by Texas A&M High Performance Research Computing.

**DATA AVAILABILITY STATEMENTS**

The datasets and code will be available at https://github.com/jungangc/PICNN-twophaseporousflow once published.

**DECLARATIONS**

The authors have no relevant financial or non-financial interests to disclose.

# REFERENCES

[1] Abidoye, L. K., Khudaida, K. J., & Das, D. B. (2015). Geological carbon sequestration in the context of two-phase flow in porous media: a review. Critical Reviews in Environmental Science and Technology, 45(11), 1105-1147.

[2] Kueper, B. H., & Frind, E. O. (1991). Two-phase flow in heterogeneous porous media: 1. Model development. Water resources research, 27(6), 1049-1057.

[3] Aziz, K. (1979). Petroleum reservoir simulation. Applied Science Publishers, 476.

[4] Chen, Z., Huan, G., & Ma, Y. (2006). Computational methods for multiphase flows in porous media. Society for Industrial and Applied Mathematics.

[5] Russell, T. F., & Wheeler, M. F. (1983). Finite element and finite difference methods for continuous flows in porous media. In The mathematics of reservoir simulation (pp. 35-106). Society for Industrial and Applied Mathematics.

[6] Tang, M., Liu, Y., & Durlofsky, L. J. (2020). A deep-learning-based surrogate model for data assimilation in dynamic subsurface flow problems. *Journal of Computational Physics*, *413*, 109456.

[7] Jin, Z. L., Liu, Y., & Durlofsky, L. J. (2020). Deep-learning-based surrogate model for reservoir simulation with time-varying well controls. *Journal of Petroleum Science and Engineering*, *192*, 107273.

[8] Coutinho, E. J. R., Dall'Aqua, M., & Gildin, E. (2021). Physics-aware deep-learning-based proxy reservoir simulation model equipped with state and well output prediction. *Frontiers in Applied Mathematics and Statistics*, *7*, 651178.

[9] Dall'Aqua, M. J., Coutinho, E. J., Gildin, E., Guo, Z., Zalavadia, H., & Sankaran, S. (2023, March). Guided Deep Learning Manifold Linearization of Porous Media Flow Equations. In *SPE Reservoir Simulation Conference*. OnePetro.

[10] Sathujoda, S. T., & Sheth, S. M. (2023). Physics-Informed Localized Learning for Advection-Diffusion-Reaction Systems. *arXiv preprint arXiv:2305.03774*.

[11] Atadeger, A., Sheth, S., Vera, G., Banerjee, R., & Onur, M. (2022, September). Deep learning-based proxy models to simulate subsurface flow of three-dimensional reservoir systems. In *ECMOR 2022* (Vol. 2022, No. 1, pp. 1-32). European Association of Geoscientists & Engineers.

[12] Mo, S., Zhu, Y., Zabaras, N., Shi, X., & Wu, J. (2019). Deep convolutional encoder-decoder networks for uncertainty quantification of dynamic multiphase flow in heterogeneous media. *Water Resources Research*, *55*(1), 703-728.

[13] Abdar, M., Pourpanah, F., Hussain, S., Rezazadegan, D., Liu, L., Ghavamzadeh, M., ... & Nahavandi, S. (2021). A review of uncertainty quantification in deep learning: Techniques, applications and challenges. *Information fusion*, *76*, 243-297.

[14] Mo, S., Zabaras, N., Shi, X., & Wu, J. (2019). Deep autoregressive neural networks for high-dimensional inverse problems in groundwater contaminant source identification. *Water Resources Research*, *55*(5), 3856-3881.

[15] Wang, N., Chang, H., & Zhang, D. (2021). Deep-learning-based inverse modeling approaches: A subsurface flow example. *Journal of Geophysical Research: Solid Earth*, *126*(2), e2020JB020549.

[16] Xiao, C., Deng, Y., & Wang, G. (2021). Deep-learning-based adjoint state method: Methodology and preliminary application to inverse modeling. *Water Resources Research*, *57*(2), e2020WR027400.

[17] Tang, H., Fu, P., Jo, H., Jiang, S., Sherman, C. S., Hamon, F., ... & Morris, J. P. (2022). Deep
22